\documentclass[11pt]{article}

\PassOptionsToPackage{table}{xcolor}
\usepackage{acl}

\usepackage{times}
\usepackage{latexsym}

\usepackage[T1]{fontenc}

\usepackage[utf8]{inputenc}

\usepackage{microtype}

\usepackage{inconsolata}

\usepackage{graphicx}

\usepackage{multirow}        
\usepackage{makecell}        
\usepackage{booktabs}        
\usepackage{algorithm}
\usepackage{algpseudocode}
\usepackage{amsmath}
\usepackage{listings}
\newcommand{\gain}[1]{\textcolor{green!60!black}{#1}}
   
\lstdefinestyle{ridgeprompt}{
  basicstyle=\ttfamily\footnotesize,
  backgroundcolor=\color{gray!6},
  frame=single,
  framerule=0.4pt,
  rulecolor=\color{black!35},
  breaklines=true,
  breakatwhitespace=true,
  columns=fullflexible,
  keepspaces=true,
  showstringspaces=false,
  xleftmargin=0.4em,
  xrightmargin=0.4em,
  aboveskip=0.6em,
  belowskip=0.6em
}

\title{RIDGE: Region-Informed Derivative-Guided Evidence Selection for Long Video Understanding}

\author{
  Shanqing Xu\textsuperscript{1}\quad
  Meng Luo\textsuperscript{2} \quad
  Mengchen Qian\textsuperscript{1} \quad
  Yuhui Gao\textsuperscript{1} \quad
  Siyue Peng\textsuperscript{1} \\
\bfseries
  Xiaohan Zhong\textsuperscript{1} \quad
  Xiaojin Zhang\textsuperscript{1} \quad
  Zhongyu Wei\textsuperscript{3} \quad
  Wei Chen\textsuperscript{1}\footnotemark[2] \quad
  Xiang Bai\textsuperscript{1} \\
  \textsuperscript{1}Huazhong University of Science and Technology \\
  \textsuperscript{2}National University of Singapore \quad \textsuperscript{3}Fudan University\\
    \textbf{Code:} \url{https://github.com/Xssq999/RIDGE}
}

\hypersetup{
  pdftitle={RIDGE: Region-Informed Derivative-Guided Evidence Selection for Long Video Understanding},
  pdfauthor={Shanqing Xu, Meng Luo, Mengchen Qian, Yuhui Gao, Siyue Peng, Xiaohan Zhong, Xiaojin Zhang, Zhongyu Wei, Wei Chen, Xiang Bai}
}

\begin{document}
\maketitle
\begingroup
\renewcommand{\thefootnote}{\fnsymbol{footnote}}
\footnotetext[0]{Email. shanqing\_xu37@hust.edu.cn}
\footnotetext[2]{Corresponding author. lemuria\_chen@hust.edu.cn}
\endgroup
\setcounter{footnote}{0}

\begin{abstract}
Long videos contain far more visual content than Large
Vision-Language Models (LVLMs) can process under a fixed visual-token
budget, making frame selection essential. Existing query-aware
selectors usually estimate frame-query relevance and build a compact
subset from high-scoring frames. Although their mechanisms differ, the
similarity sequence is still often treated primarily as values to rank
or sample from, rather than as an ordered signal whose shape reflects
how query-relevant evidence emerges, peaks, and fades over time. This
can obscure frames that explain, contextualize, or follow an event,
because such evidence may lie on the rising or falling sides of a
nearby relevance peak and receive lower absolute scores. We propose
RIDGE, a frame selection framework that reads the frame-query
similarity curve as a temporal signal. By using local changes and
curvature, RIDGE partitions the timeline into structural regions
and applies region-specific selection to preserve event cores,
transitions, buildup, aftermath, and contextual frames under a fixed
budget. It is a lightweight post-processing step on precomputed
frame-query scores and requires neither training nor iterative LVLM
calls. Across four long-video benchmarks and three backbones,
RIDGE achieves the best performance in most settings and remains
competitive in the others.
\end{abstract}

\begin{figure}[t]
  \includegraphics[width=\columnwidth]{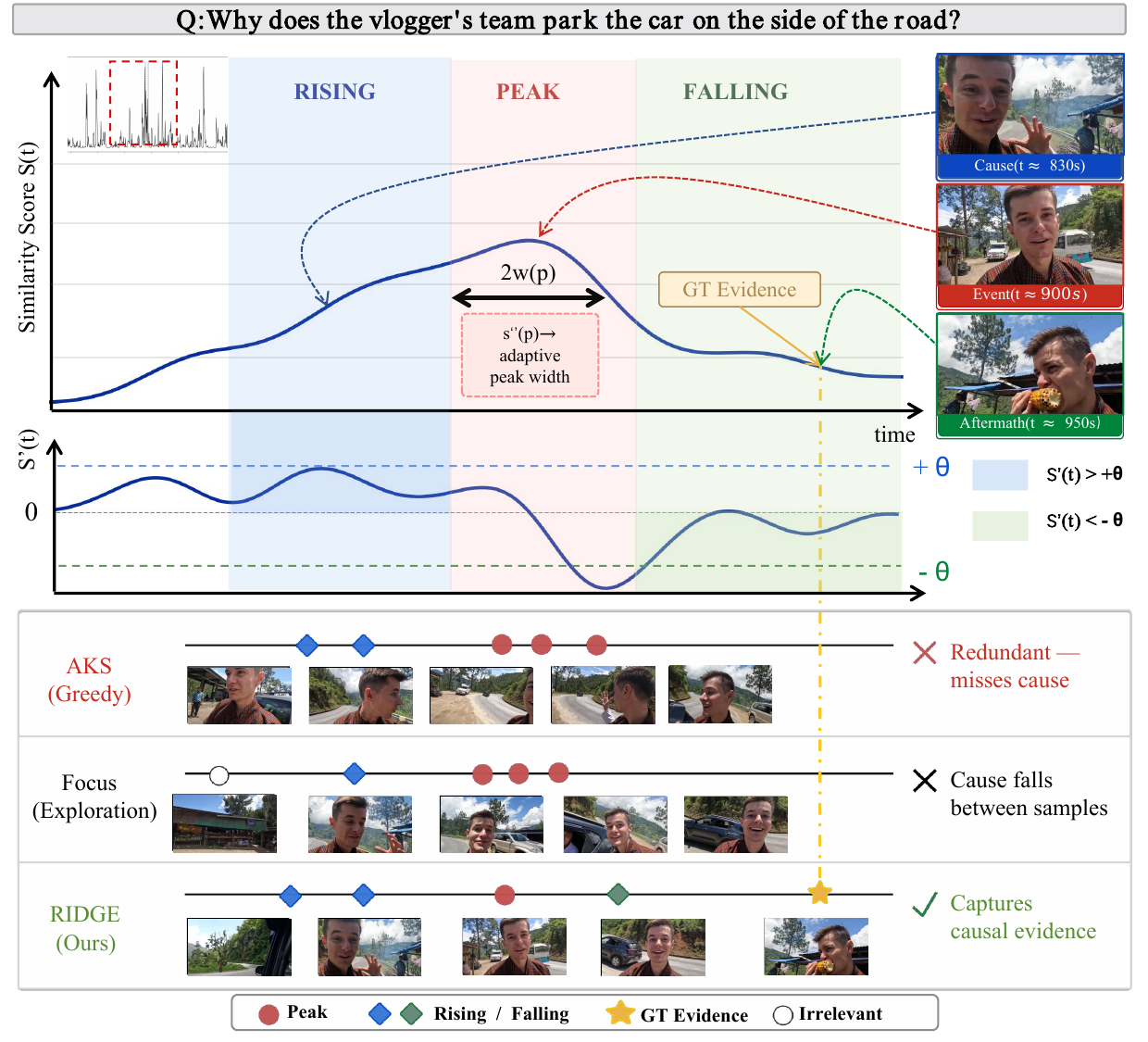}
  \caption{The question requires evidence from the cause, event, and aftermath of a long video. RIDGE samples the rising--peak--falling structure of the query-relevance curve to recover this distributed evidence.}
  \label{fig:comparative_methods}
\end{figure}

\section{Introduction}

Large Vision-Language Models (LVLMs) have shown strong performance on video captioning, retrieval, and question answering \citep{luo2024panosent,maaz2024video}. Long-video understanding, however, remains challenging \citep{nguyen2024video}: even a one-hour video sampled at 1 FPS yields thousands of frames and millions of visual tokens. Although modern LVLMs can process increasingly long multimodal contexts, feeding such inputs in full is still computationally costly and often inefficient \citep{luo2026unveiling,tang2025adaptive}, especially when the evidence relevant to a user query lies in only a small portion of the video. Frame selection \citep{liang2024keyvideollm,tang2025adaptive} therefore remains a practical way to reduce visual redundancy and focus computation on query-relevant evidence. Given a video of \(T\) frames and a textual query, the selector must choose a small subset of \(N \ll T\) frames that preserves the evidence needed by the downstream model \citep{hu2025m, luo2026dr}. Consequently, selecting only the highest-scoring frames can miss the buildup and aftermath evidence needed to distinguish an event from its cause or consequence.

A common query-aware approach is to compute a frame-query similarity score \(s_t=\mathrm{sim}(Q,F_t)\) and select frames from this score sequence \citep{radford2021learning, liang2024keyvideollm,hu2025m}. Recent methods enrich this pipeline with temporal grouping \citep{wang2025videotree}, diversity objectives \citep{liu2025bolt,hu2025m}, adaptive search \citep{tang2025adaptive,ye2025re}, or iterative reasoning \citep{wang2025videotree,ye2025re}. However, they still largely read the similarity curve by its magnitude: high-scoring frames are treated as the strongest evidence, while temporal structure is used mainly for coverage or search \citep{liu2025bolt,tang2025adaptive,ye2025re}. This can be limiting when the answer depends not only on the most explicit moment, but also on how that moment is reached or what changes around it.


We therefore take a different view: the frame-query similarity sequence is not merely a list of relevance scores, but a query-conditioned temporal signal. Its local shape can reveal whether the video is approaching a relevant event, staying at its core, moving away from it, crossing a sharp transition, or providing broader context. Based on this view, we propose \textbf{RIDGE}, a training-free query-aware frame selector that reads the similarity curve by both height and shape.

Given precomputed frame-query scores, RIDGE first segments the smoothed similarity curve into structural regions using local slope and curvature. It then allocates the frame budget across regions according to question preference and applies region-matched selection rules. As a result, highly relevant event cores are preserved without consuming the entire budget, leaving room for buildup, aftermath, transitions, and context that may be necessary for answering long-video questions. RIDGE is lightweight, backbone-agnostic, and does not require iterative calls to the downstream LVLM.

Our \textbf{main contributions} are threefold: 1) We reinterpret query-aware keyframe selection as reading a query-conditioned temporal signal, showing that the local shape of the similarity curve encodes different evidence roles beyond score magnitude alone; 2) We propose RIDGE, a lightweight training-free selector that segments the similarity curve into structural regions and performs question-aware, region-specific frame selection to preserve both event cores and surrounding temporal evidence; 3) We validate RIDGE on four long-video benchmarks and three LVLM backbones, demonstrating consistent gains over uniform sampling and competitive or superior performance to stronger adaptive and LVLM-assisted baselines.




\begin{figure*}[t]
    \centering
    \includegraphics[width=1.0\textwidth]{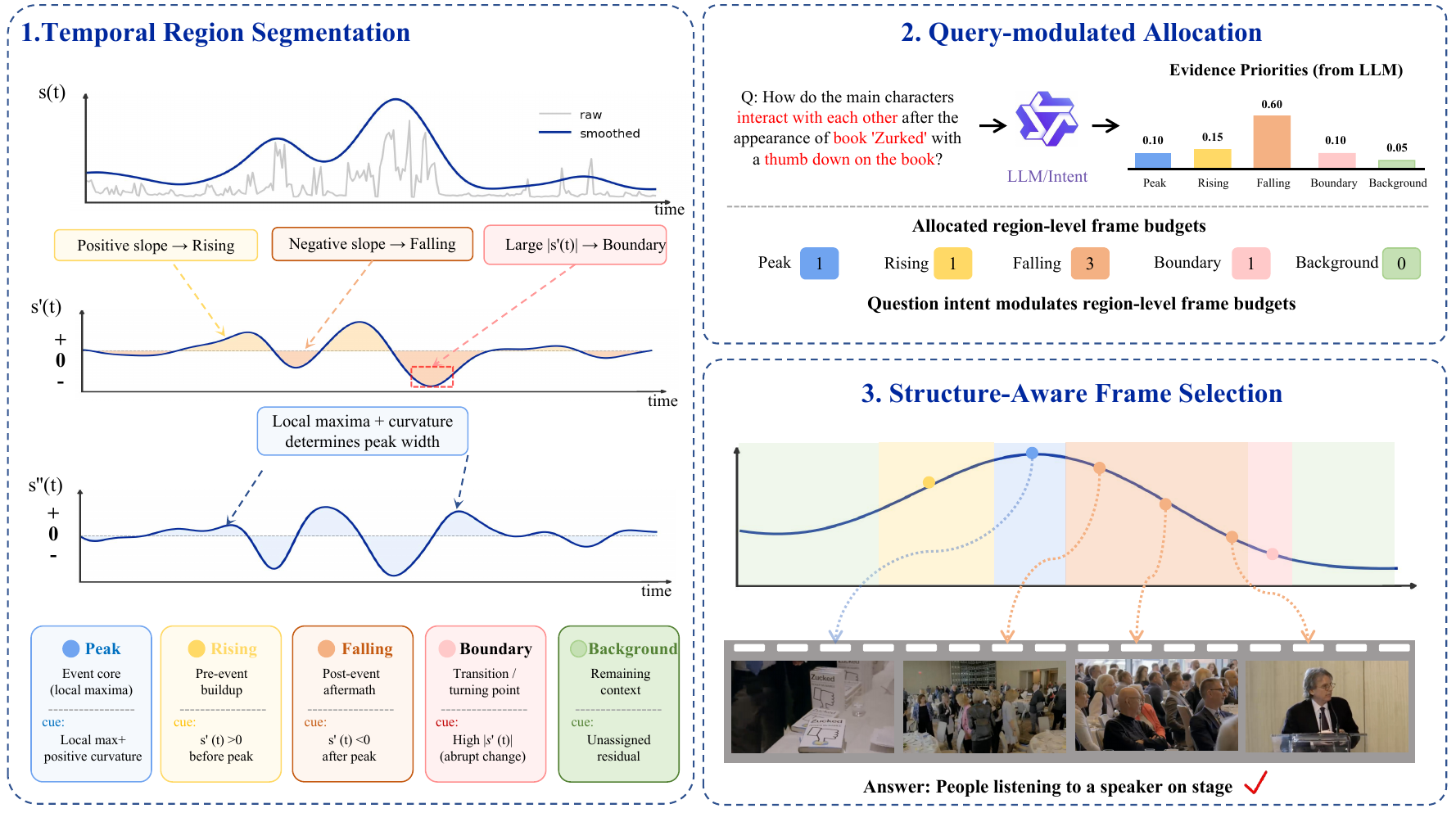}
    \caption{Overview of RIDGE. Given query-frame relevance scores, RIDGE segments the temporal signal into evidence roles, allocates the frame budget with question-aware preferences, and selects role-specific frames for LVLM inference.}
    \label{fig:method}
\end{figure*}

\section{Related Work}

\paragraph{LVLMs for Long Video Understanding.}
LVLMs have rapidly evolved from
image-centric multimodal systems to models capable of processing video
inputs \citep{tang2025video,fei2025current}. For long-video understanding, recent models mainly improve the
way visual evidence is represented once it enters the model \citep{maaz2024video}. Some
systems improve the fidelity of visual inputs through dynamic
resolution, time-aware encoding, unified image-video training, or larger
visual backbones, as in Qwen2.5-VL \citep{bai2025qwen25vltechnicalreport}, LLaVA-OneVision \citep{li2024llava}, and InternVL-3 \citep{zhu2025internvl3}.
Others reduce the burden of long inputs through temporal aggregation,
visual token compression, or memory-style mechanisms, as in Video-LLaMA \citep{zhang2023video},
LLaVA-Video \citep{zhang2024llava}, and LongVU \citep{shen2024longvu}. These advances strengthen the downstream
reasoner \citep{shu2025video,zhang2024long}, but they still operate on a finite visual input. In practical
long-video QA, the model usually sees a sampled or compressed view of
the original video \citep{song2024moviechat}. Architectural scaling can improve how the model uses
visible evidence, while frame selection determines which evidence becomes
visible before reasoning begins.

\paragraph{Keyframe Selection.}
Keyframe selection controls video input view by choosing a compact set of frames for downstream reasoning. Query-agnostic strategies, such as
uniform sampling or shot-based summarization, provide broad temporal
coverage but cannot adapt to the evidence required by a particular
question. Query-aware methods therefore build a relevance signal between
the query and candidate frames or clips. The simplest form is top-\(k\)
selection, while stronger variants use the same signal together with
temporal structure or selection constraints. AKS \citep{tang2025adaptive}, for example, balances
relevance and coverage through recursive temporal partitioning; MDP\(^3\)
\citep{sun2025mdp3} adds diversity so that selected frames are less redundant; and stochastic
selectors such as BOLT \citep{liu2025bolt} and Q-Frame \citep{zhang2025q} reduce the brittleness of always
choosing the highest-scoring frames. Another family allocates computation
more actively: FOCUS \citep{zhu2025focus} searches over clips under a fixed budget, while
A.I.R. \citep{zou2025air} and Frame-Voyager \citep{yu2025frame} use LVLM feedback or supervision to refine the
candidate evidence.
These methods improve keyframe selection from different angles:
coverage, diversity, stochasticity, adaptive search, or model-guided
refinement. From the perspective of RIDGE, what remains
underexplored is the internal temporal shape of the relevance signal.

\section{Method}

\subsection{Overview}
Let \(V=\{F_1,\ldots,F_T\}\) be the candidate frames of a video and
\(Q\) be the question. A lightweight vision-language scorer produces a
query-frame relevance sequence
\[
    s_t=\mathrm{sim}(Q,F_t), \quad t=1,\ldots,T .
\]
Given a frame budget \(N\), RIDGE selects
\(\min(N,T)\) frame indices and passes the corresponding frames to the
downstream LVLM in temporal order. RIDGE also uses a compact
evidence-preference vector
\[
\begin{aligned}
\mathbf{w}=(&w_{\mathrm{peak}},w_{\mathrm{slope}},w_{\mathrm{rise}},\\
&w_{\mathrm{fall}},w_{\mathrm{boundary}},w_{\mathrm{context}}),
\end{aligned}
\]
which is either fixed or generated once from the question. 

In many video questions, the answer is not confined to a single instant: the lead-up to an event, its immediate aftermath, or the moment where relevance changes abruptly may be just as informative as the highest-scoring frame itself. Our method RIDGE is built around a simple premise: \textbf{\emph{a relevance curve should not be read only by its height}}. Its shape also tells us whether the video is moving toward a relevant event, staying at the event core, moving away from it, or crossing an abrupt transition.

This suggests that frame selection should answer three distinct questions: 1) \textbf{\emph{what temporal role each frame plays}}; 2) \textbf{\emph{how much budget each role deserves for the current question}} and 3) \textbf{\emph{which representative frames best express that role}}. RIDGE therefore separates frame selection into three decisions. First, it converts the one-dimensional relevance curve into a temporal role map
with five regions:
\[
\begin{aligned}
\mathcal{Y}=\{&\textsc{peak},\textsc{rising},\textsc{falling},\\
&\textsc{boundary},\textsc{background}\}.
\end{aligned}
\]
\textsc{Peak} frames capture event cores, \textsc{rising} frames capture
pre-event buildup, \textsc{falling} frames capture aftermath,
\textsc{boundary} frames capture abrupt changes, and
\textsc{background} frames preserve remaining context. Second, RIDGE
allocates the frame budget across these roles according to the evidence
needed by the question. Third, it selects frames inside each region with
a rule matched to that region's temporal function.

\subsection{Temporal Region Segmentation}
RIDGE starts by making the score sequence stable enough for temporal
analysis. It min--max normalizes \(s_t\) when the score range is
non-zero, smooths the normalized sequence with a Gaussian kernel, and
computes its first- and second-order differences:
\[
    \bar{s}_t=(G_\sigma * \hat{s})(t),\qquad
    s'_t=\nabla\bar{s}_t,\qquad s''_t=\nabla s'_t .
\]

\begin{algorithm}[H]
\footnotesize
\caption{RIDGE: Curve-guided and Query-aware Frame Selection}
\label{alg:ridge}
\begin{algorithmic}[1]
  \Require Scores \(s_{1:T}\), budget \(N\), preference vector \(\mathbf{w}\), hyperparameters \(\Theta\)
  \Ensure Selected indices \(\mathcal{I}\) in temporal order
  \If{\(T \leq N\)} \State \Return \(\{1,\ldots,T\}\) \EndIf
  \State Normalize and smooth scores; compute \(s'=\nabla\bar{s}\) and \(s''=\nabla s'\)
  \State Compute coverage intent \(c\) from \(\mathbf{w}\); initialize labels as \textsc{background}
  \Statex
  \State \textbf{Stage 1: curve-to-region segmentation}
  \State Set \(\mathcal{P}\) to peaks on \(\bar{s}\); if empty, use \(\{\arg\max_t\bar{s}_t\}\)
  \For{each \(p\in\mathcal{P}\)}
    \State Label a curvature-adaptive window around \(p\) as \textsc{peak}
    \State Extend to adjacent positive/negative slope spans as \textsc{rising}/\textsc{falling}, modulated by \(c\)
  \EndFor
  \State Label residual high-\(|s'|\) frames as \textsc{boundary}; keep the rest as \textsc{background}
  \Statex
  \State \textbf{Stage 2: question-aware budget allocation}
  \State Map \(\mathbf{w}\) to region priorities \(W_r\)
  \State Allocate region budgets \(n_r\) proportionally, with capacity caps and leftover redistribution
  \Statex
  \State \textbf{Stage 3: region-specific frame selection}
  \State \(\mathcal{I}\gets\emptyset\)
  \For{each region \(r\) with \(n_r>0\)}
    \If{\(r\in\{\textsc{rising},\textsc{falling}\}\) and \(n_r\geq3\) and \(|\mathcal{R}_r|\geq3\)}
      \State \(\mathcal{I}\gets\mathcal{I}\cup\textsc{BinSelect}(\mathcal{R}_r,\bar{s},n_r)\)
    \ElsIf{\(r=\textsc{boundary}\)}
      \State \(\mathcal{I}\gets\mathcal{I}\cup\textsc{RankSelect}(\mathcal{R}_r,|s'|,n_r)\)
    \Else
      \State \(\mathcal{I}\gets\mathcal{I}\cup\textsc{RankSelect}(\mathcal{R}_r,\bar{s},n_r)\)
    \EndIf
  \EndFor
  \State Deduplicate \(\mathcal{I}\); if \(|\mathcal{I}|<N\), backfill unselected frames by descending \(\bar{s}_t\)
  \State \Return first \(\min(N,T)\) temporally sorted indices in \(\mathcal{I}\)
\end{algorithmic}
\end{algorithm}

The smoothed score \(\bar{s}_t\) estimates how relevant a frame is, the
slope \(s'_t\) describes whether relevance is increasing or decreasing,
and the curvature \(s''_t\) helps determine whether a relevance peak is
sharp or temporally extended. This turns the score sequence from isolated framewise values into a local temporal profile whose level, slope, and sharpness can be read as complementary evidence cues.

\paragraph{Event cores.}
RIDGE first finds local maxima on \(\bar{s}\) and treats them as
candidate event cores. Around each peak \(p\), it assigns a
curvature-adaptive \textsc{peak} window. The half-width is inversely
related to the relative curvature
\[
    r_p=\frac{|s''_p|}{\mathrm{median}(|s''|)+\epsilon}.
\]
RIDGE retains all local maxima that pass the prominence threshold; the
global maximum is used only as a fallback when no peak is detected.
Each detected peak receives its own \textsc{peak} window and adjacent
\textsc{rising}/\textsc{falling} extensions, although RIDGE does not
impose a hard frame quota for every peak. Appendix~\ref{sec:appendix_multi_peak}
provides a quantitative analysis of this multi-peak behavior.
Sharp peaks receive compact windows because the relevant evidence is
temporally concentrated; flatter peaks receive wider windows because the
event remains visually relevant for longer. This fallback ensures that
every video has at least one event core, even in weakly structured or
noisy cases, preventing the method from degenerating into pure background
selection.

\paragraph{Buildup and aftermath.}
After locating event cores, RIDGE searches outward from each peak to
recover neighboring temporal phases. Frames on the left are labeled
\textsc{rising} while the slope remains sufficiently positive, and
frames on the right are labeled \textsc{falling} while the slope remains
sufficiently negative.

The extension is modulated by the question. Given the evidence
preference vector \(\mathbf{w}\), RIDGE computes a coverage intent
\[
    c=
    \frac{w_{\mathrm{rise}}+w_{\mathrm{fall}}+w_{\mathrm{context}}}
    {w_{\mathrm{peak}}+w_{\mathrm{rise}}+
     w_{\mathrm{fall}}+w_{\mathrm{context}}+\epsilon}.
\]
A larger \(c\) means the question is expected to require broader
temporal evidence, so the rising/falling search uses a looser slope
threshold and a longer maximum extension. A smaller \(c\) keeps the regions tighter around the event core. Intuitively, some questions behave like snapshot queries, while others can only be answered by tracing a short temporal development. The exact thresholding rule is
listed in Appendix~\ref{sec:appendix_ridge_details}.

\paragraph{Transitions and context.}
Once peak, rising, and falling regions have been assigned, RIDGE looks
only at the residual background frames. Residual frames with unusually
large \(|s'_t|\) are labeled as \textsc{boundary}, capturing abrupt
changes in the query-relevance trajectory that may correspond to scene
cuts or action switches. All remaining frames retain the
\textsc{background} label.

\subsection{Question-aware Budget Allocation}

The region map identifies the temporal role of each frame, while
\(\mathbf{w}\) specifies which evidence types are likely to be useful
for the question. A question may focus mainly on the event core or
require a broader account of how the event is prepared, interrupted, or
resolved. The entries of \(\mathbf{w}\) are not frame scores; they
describe what kinds of visual evidence are likely to support the answer.

RIDGE maps the six evidence preferences to five region priorities by
letting \textsc{peak}, \textsc{rising}, \textsc{falling}, and
\textsc{background} inherit their corresponding weights, while setting
\(W_{\textsc{boundary}}=w_{\mathrm{boundary}}+w_{\mathrm{slope}}\).
Here, \(w_{\mathrm{rise}}\) and \(w_{\mathrm{fall}}\) encode directional
phase preferences, whereas \(w_{\mathrm{slope}}\) encodes rapid relevance
change regardless of direction. We route \(w_{\mathrm{slope}}\) to
\textsc{boundary} because \textsc{boundary} contains residual high-\(|s'|\)
frames and is the only region whose internal selection rule ranks by
\(|s'|\). Routing it additionally to \textsc{rising} and
\textsc{falling} could double-count slope preference in phases that
already have dedicated weights.

For each non-empty region, RIDGE assigns a proportional share of the
frame budget according to \(W_r\), caps the share by the number of
available frames in that region, and redistributes any unused budget to
the remaining available regions. This keeps the allocation faithful to
the question while respecting the actual structure discovered in the
video.

\subsection{Role-specific Frame Selection}
The final stage chooses concrete frames from each allocated region.
RIDGE uses score ranking where high absolute relevance is the right
criterion, and temporal coverage where the region's value lies in its
evolution over time. 

For \textsc{peak}, RIDGE ranks frames by \(\bar{s}_t\), keeping the
clearest views of the event core. \textsc{Background} uses the same
ranking so that contextual frames remain query-relevant. For
\textsc{boundary}, RIDGE ranks by \(|s'_t|\), because what matters there is not how relevant a frame already is, but how strongly it marks a temporal change in relevance.
For \textsc{rising} and \textsc{falling}, RIDGE uses temporal sub-bin
selection when the region and its budget are large enough: it splits the
region into equal temporal bins and selects the highest-\(\bar{s}_t\)
frame from each bin. The goal is not only to keep relevant frames, but to preserve the temporal spread of evidence inside a phase. This avoids collapsing all buildup or aftermath
frames near the adjacent peak. When a region is too small for meaningful
binning, RIDGE falls back to score ranking.

Frames selected from all regions are merged, deduplicated, and sorted by
time. If the selected set is still smaller than the target budget, RIDGE
backfills from the remaining frames by descending \(\bar{s}_t\). This preserves the role-aware structure whenever possible, while still using the available budget on frames that remain broadly relevant to the question.


\section{Experiments}

\begin{table*}[t]
\centering
\caption{Comparison of LVLMs and frame selection methods on Video-MME, LongVideoBench, MLVU, and LVBench. We report accuracy scores (\%). \textbf{Bold} and \underline{underline} indicate the best and second-best performance within each model group.}
\label{tab:main_results}
\setlength{\tabcolsep}{5pt}

\newcommand{\up}[1]{{\color{green!60!black}\scriptsize$\uparrow$\!#1}}
\newcommand{\down}[1]{{\color{red!70!black}\scriptsize$\downarrow$\!#1}}

\newcommand{\score}[2]{\makebox[3.8em][r]{#1}\,\makebox[2.4em][l]{#2}}
\newcommand{\base}[1]{\textcolor{gray}{\score{#1}{}}}
\newcommand{\miss}{\score{--}{}}

\newcommand{\hscore}[2]{\cellcolor{yellow!15}\score{#1}{#2}}
\newcommand{\htext}[1]{\cellcolor{yellow!15}#1}

\resizebox{1.0\textwidth}{!}{%
\begin{tabular}{l|l|c|cccc}
\toprule
\textbf{Model} & \textbf{Method} & \textbf{Frame}
  & \textbf{Video-MME}
  & \textbf{LongVideoBench}
  & \textbf{MLVU}
  & \textbf{LVBench} \\
\midrule

\multirow{6}{*}{LLaVA-OV-7B}
& \color{gray}{Baseline} & \color{gray}{32}
  & \base{56.5} & \base{54.8} & \base{63.1} & \base{40.4} \\
& AKS & 32
  & \score{58.4}{\up{1.9}}
  & \score{59.3}{\up{4.5}}
  & \score{67.5}{\up{4.4}}
  & \score{47.3}{\up{6.9}}\\
& MDP$^3$ & 32
  & \score{\underline{58.5}}{\up{2.0}}
  & \score{59.0}{\up{4.2}}
  & \score{\textbf{68.2}}{\up{5.1}}
  & \score{41.6}{\up{1.2}}\\
& BOLT & 32
  & \score{57.9}{\up{1.4}}
  & \score{59.6}{\up{4.8}}
  & \score{66.8}{\up{3.7}}
  & \score{38.2}{\down{2.2}} \\
& FOCUS & 32
  & \score{58.3}{\up{1.8}}
  & \score{\textbf{60.7}}{\up{5.9}}
  & \score{67.1}{\up{4.0}}
  & \score{\underline{47.5}}{\up{7.1}} \\
& \htext{RIDGE} & \htext{32}
  & \hscore{\textbf{59.6}}{\up{3.1}}
  & \hscore{\underline{60.1}}{\up{5.3}}
  & \hscore{\underline{68.0}}{\up{4.9}}
  & \hscore{\textbf{48.9}}{\up{8.5}} \\

  \midrule

\multirow{6}{*}{InternVL-3-8B}
& \color{gray}{Baseline} & \color{gray}{32}
  & \base{65.6} & \base{58.5} & \base{68.4} & \base{43.5} \\
& AKS & 32
  & \score{66.3}{\up{0.7}}
  & \score{\underline{61.5}}{\up{3.0}}
  & \score{\underline{74.2}}{\up{5.8}} 
  & \score{52.4}{\up{8.9}} \\
& MDP$^3$ & 32
  & \score{\textbf{66.8}}{\up{1.2}}
  & \score{60.9}{\up{2.4}}
  & \score{73.9}{\up{5.5}} 
  & \score{47.3}{\up{3.8}} \\
& BOLT & 32
  & \score{66.1}{\up{0.5}}
  & \score{59.2}{\up{0.7}}
  & \score{70.8}{\up{2.4}}
  & \score{44.2}{\up{0.7}} \\
& FOCUS & 32
  & \score{64.3}{\down{1.3}}
  & \score{61.0}{\up{2.5}}
  & \score{72.0}{\up{3.6}}
  & \score{\underline{53.2}}{\up{9.7}} \\
& \htext{RIDGE} & \htext{32}
  & \hscore{\underline{66.6}}{\up{1.0}}
  & \hscore{\textbf{61.8}}{\up{3.3}}
  & \hscore{\textbf{74.6}}{\up{6.2}}
  & \hscore{\textbf{54.5}}{\up{11.0}} \\

\midrule

\multirow{6}{*}{Qwen2.5-VL-7B}
& \color{gray}{Baseline} & \color{gray}{32}
  & \base{61.2} & \base{58.9} & \base{59.7} & \base{38.5} \\
& AKS & 32
  & \score{63.1}{\up{1.9}}
  & \score{\underline{63.2}}{\up{4.3}}
  & \score{67.2}{\up{7.5}}
  & \score{43.4}{\up{4.9}} \\
& MDP$^3$ & 32
  & \score{\textbf{63.8}}{\up{2.6}}
  & \score{60.0}{\up{1.1}}
  & \score{66.2}{\up{6.5}}
  & \score{41.3}{\up{2.8}} \\
& BOLT & 32
  & \score{63.2}{\up{2.0}}
  & \score{60.0}{\up{1.1}}
  & \score{\underline{68.4}}{\up{8.7}}
  & \score{39.3}{\up{0.8}} \\
& FOCUS & 32
  & \score{63.4}{\up{2.2}}
  & \score{61.0}{\up{2.1}}
  & \score{68.2}{\up{8.5}}
  & \score{\underline{46.3}}{\up{7.8}} \\
& \htext{RIDGE} & \htext{32}
  & \hscore{\underline{63.7}}{\up{2.5}}
  & \hscore{\textbf{65.4}}{\up{6.5}}
  & \hscore{\textbf{69.6}}{\up{9.9}}
  & \hscore{\textbf{50.9}}{\up{12.4}} \\

\bottomrule
\end{tabular}%
    }
\end{table*}

\subsection{Experimental Setup}

\noindent\textbf{Datasets and Baselines.}
We evaluate RIDGE on four widely used long-video understanding benchmarks.
Video-MME \citep{fu2025video} consists of 900 videos covering short, medium, and long durations and is annotated with 2{,}700 multiple-choice QA pairs. We report results without subtitles.
MLVU \citep{zhou2025mlvu} contains 1{,}730 long videos with an average duration of 12 minutes and covers nine task types. We report M-Avg on the dev set.
LongVideoBench (LVB)
\citep{wu2024longvideobench} provides 6{,}678 QA pairs over 3{,}763 videos ranging from 8 seconds up to 1 hour, and emphasizes long-context referring reasoning. We report results on the validation set.
LVBench \citep{wang2025lvbench} contains 103 hour-long videos and 1{,}549 QA pairs requiring long-horizon temporal reasoning.

We compare RIDGE with uniform sampling, score-ranking or adaptive selectors including AKS~\citep{tang2025adaptive}, BOLT~\citep{liu2025bolt}, and FOCUS~\citep{zhu2025focus}, and the diversity-aware selector MDP$^3$~\citep{sun2025mdp3}. All baseline and competing-method results in
Table~\ref{tab:main_results} were obtained by rerunning the official
implementations under our unified evaluation pipeline, using each
method's default configuration. The main comparison retains each method's default scorer: AKS and FOCUS use BLIP-ITM, BOLT uses CLIP, MDP$^3$ uses SigLIP, and RIDGE uses BLIP-ITM; therefore, Table~\ref{tab:main_results} is not a shared-scorer comparison. Table~\ref{tab:matched_scorer} reports matched-scorer controls that isolate the selection policy.

\noindent\textbf{Implementation Details.}
We evaluate RIDGE on three representative LVLM backbones: Qwen2.5-VL-7B \citep{bai2025qwen25vltechnicalreport}, InternVL-3-8B \citep{zhu2025internvl3}, and LLaVA-OV-7B \citep{li2024llava}, feeding the selected frames in temporal order and following the official inference settings of each backbone. Unless otherwise stated, query--frame similarities $\{s_t\}$ are computed with BLIP-ITM \citep{li2022blip}. 
RIDGE uses a 6-dimensional evidence-preference vector \(\mathbf{w}\), instantiated either as a fixed default vector or as question-specific weights generated once by Qwen3-8B \citep{yang2025qwen3technicalreport}. We ablate Qwen3-3B and Llama-3.1-8B \citep{grattafiori2024llama3herdmodels} as alternative weight generators, and CLIP \citep{radford2021learning}, BLIP-2 \citep{li2023blip}, SigLIP \citep{zhai2023sigmoid}, and BLIP-ITM \citep{li2022blip} as scoring models. The default frame budget is $N{=}32$. RIDGE's frame-selection stage is deterministic. We use one shared RIDGE hyperparameter configuration across benchmarks, LVLM backbones, frame budgets, model scales, and scorers; the sensitivity sweeps were conducted after fixing the default and are used only as robustness analyses. RIDGE itself is a training-free, plug-in selector that runs as a lightweight CPU post-processing step and incurs no extra cost during LVLM inference. In our experiments, LVLM inference is performed on four NVIDIA A100 GPUs, and the pipeline is also deployed and applied on Ascend graphics cards. We report accuracy measured by
the LMMs-Eval toolkit \citep{zhang2025lmms}. Full hyperparameter settings are deferred to the appendix.

\subsection{Comparison with SOTA Methods}

Table~\ref{tab:main_results} reports our main comparison on Video-MME, LVB, MLVU, and LVBench across three LVLM backbones. RIDGE consistently improves over uniform sampling at the same frame budget and remains competitive with stronger adaptive and diversity-aware selectors. On Qwen2.5-VL-7B, RIDGE improves the uniform baseline by 2.5 percentage points on Video-MME, 6.5 on LVB, 9.9 on MLVU, and 12.4 on LVBench. The same pattern holds for InternVL-3-8B and LLaVA-OV-7B, showing that the gains are not tied to one downstream LVLM. The improvements are especially large on MLVU and LVBench, which contain more long-horizon and multi-event reasoning cases, supporting our hypothesis that temporal-shape-aware selection is particularly beneficial when evidence is distributed across event buildup, core moments, and aftermath. The comparatively modest gain on Video-MME is expected: its benchmark is dominated by short- and medium-length clips whose key information is often concentrated in a few visually distinct shots, leaving limited room for a smarter selection policy to add value.

\noindent\textbf{Matched-Scorer Comparison.}
The main comparison in Table~\ref{tab:main_results} retains the scorer
used by each method's default configuration. To separate scorer choice
from the frame-selection policy, Table~\ref{tab:matched_scorer} compares
RIDGE with BOLT and MDP$^3$ under their respective scorers. All results
use Qwen2.5-VL-7B with a frame budget of $N{=}32$.

\begin{table*}[t]
\centering
\small
\caption{Matched-scorer comparison under Qwen2.5-VL-7B with
$N{=}32$. Accuracy (\%) is reported. Within each scorer block, the
baseline and RIDGE use the same frame--query scorer.}
\label{tab:matched_scorer}
\setlength{\tabcolsep}{9pt}
\begin{tabular}{llccccc}
\toprule
Scorer & Method & Video-MME & LVB & MLVU & LVBench & Avg. \\
\midrule
\multirow{2}{*}{CLIP}
& BOLT  & 63.2 & 60.0 & 68.4 & 39.3 & 57.7 \\
& RIDGE & 64.3 & 62.0 & 69.7 & 48.9 & 61.2 \\
\midrule
\multirow{2}{*}{SigLIP}
& MDP$^3$ & 63.8 & 60.0 & 66.2 & 41.3 & 57.8 \\
& RIDGE   & 62.5 & 64.7 & 71.2 & 47.8 & 61.6 \\
\bottomrule
\end{tabular}
\end{table*}

With CLIP scores, RIDGE improves the four-benchmark average over BOLT
by 3.5 percentage points. With SigLIP scores, it improves the average
over MDP$^3$ by 3.8 percentage points. RIDGE is higher in seven of the
eight matched comparisons; the exception is Video-MME with SigLIP,
where it is 1.3 percentage points below MDP$^3$. These controls show
that the gains in Table~\ref{tab:main_results} are not attributable to
scorer choice alone.

\noindent\textbf{Effect of Question-aware Weighting.}
To isolate the contribution of question-specific weight generation, we
compare the default Qwen3-8B generator with Qwen3-0.6B and a fixed weight
vector shared by all questions. Table~\ref{tab:weight_source_comparison}
also includes the LLM-free baselines FOCUS and AKS for reference. All
results use Qwen2.5-VL-7B with a frame budget of $N{=}32$.

\begin{table*}[t]
\centering
\small
\caption{Effect of question-aware weight generation across four
benchmarks. Accuracy (\%) is reported. Avg. is the average over the four
benchmarks, and $\Delta$ is measured relative to Qwen3-8B.}
\label{tab:weight_source_comparison}
\setlength{\tabcolsep}{7pt}
\begin{tabular}{lcccccc}
\toprule
Weight Source / Method & Video-MME & LVB & MLVU & LVBench & Avg. & $\Delta$ \\
\midrule
Qwen3-8B (default)    & 63.7 & 65.4 & 69.6 & 50.9 & 62.4 & 0.0 \\
Qwen3-0.6B            & 63.2 & 64.9 & 69.9 & 49.7 & 61.9 & -0.5 \\
RIDGE (fixed weights) & 62.9 & 63.8 & 69.4 & 48.3 & 61.1 & -1.3 \\
FOCUS                 & 63.4 & 61.0 & 68.2 & 46.3 & 59.7 & -2.7 \\
AKS                   & 63.1 & 63.2 & 67.2 & 43.4 & 59.2 & -3.2 \\
\bottomrule
\end{tabular}
\end{table*}

RIDGE with fixed weights achieves a four-benchmark average of 61.1,
outperforming the LLM-free baselines FOCUS and AKS by 1.4 and 1.9
percentage points, respectively. This result shows that temporal region
modeling and region-specific selection remain effective without an
auxiliary weight-generation LLM. Question-specific weights generated by
Qwen3-8B further improve the average to 62.4, while Qwen3-0.6B retains
most of this benefit with a substantially smaller auxiliary model.

\subsection{Robustness Analysis}
\label{sec:robustness}

We stress-test RIDGE along five axes: the frame budget, the scale of the downstream LVLM, the choice of frame--query scoring model, the LLM used for weight generation, and the smoothing scale $\sigma$. Unless otherwise stated, experiments use LVB with Qwen2.5-VL-7B as the default backbone and BLIP-ITM as the default scoring model; when the LLM weight mode is used, Qwen3-8B is the default weight generator.

\begin{figure}[t]
    \centering
    \includegraphics[width=1.0\columnwidth]{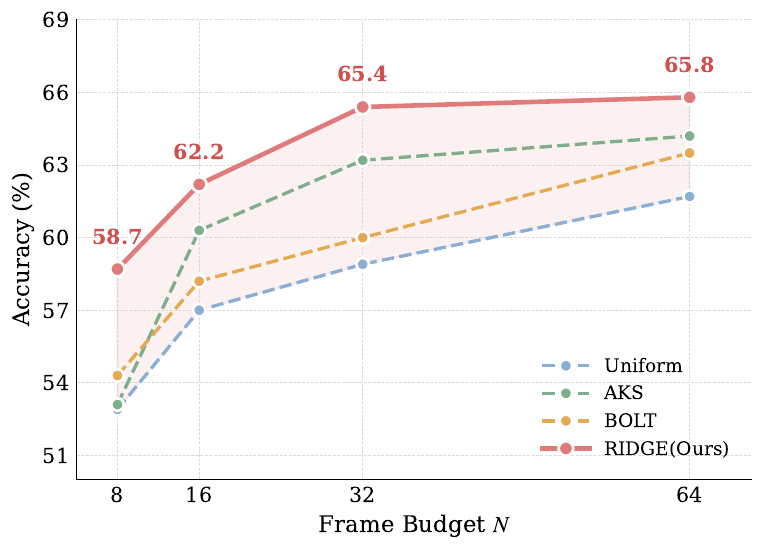}
    \caption{Accuracy on LongVideoBench under different frame budgets.}
    \label{fig:frame_budget}
\end{figure}

\noindent\textbf{Robustness to Frame Budget.}
Figure~\ref{fig:frame_budget} reports accuracy on LVB at $N \in \{8, 16, 32, 64\}$. RIDGE improves over uniform sampling across all budgets. The gain is positive even at $N{=}64$, indicating that the method is not only useful under extremely tight frame budgets. The margin over AKS is largest at smaller budgets, consistent with the intuition that explicit modeling of rising and falling regions matters most when redundant peak-centered frames are costly.

\noindent\textbf{Robustness to LVLM Scale.}
We study whether the benefit of RIDGE persists as the downstream LVLM scales up. Table~\ref{tab:model_scale} compares uniform sampling and RIDGE on Qwen2.5-VL-3B/7B/32B/72B at $N{=}32$. RIDGE improves performance at every scale. While the absolute gain remains relatively stable across model sizes, even the 72B model still benefits by 6.3 percentage points, indicating that better frame selection is complementary to LVLM scaling rather than absorbed by larger models.

\begin{figure}[t]
    \centering
    \includegraphics[width=1.0\columnwidth]{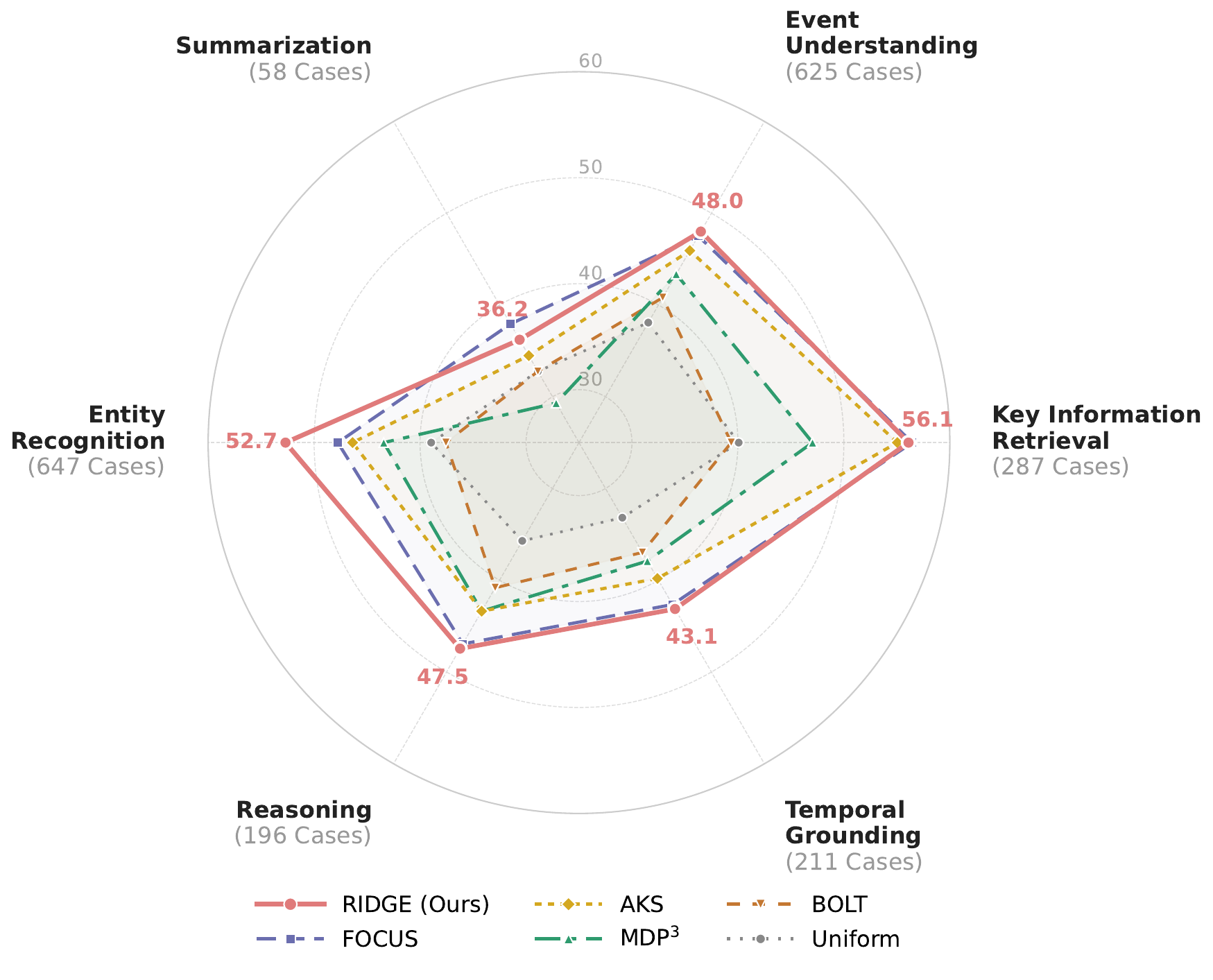}
    \caption{Per-question-type accuracy on LVBench.}
    \label{fig:question_type}
\end{figure}

\begin{table}[t]
\centering
\small
\caption{Robustness across LVLM scales on LongVideoBench.}
\label{tab:model_scale}
\setlength{\tabcolsep}{6pt}
\begin{tabular}{lccc}
\toprule
Model Scale & Uniform & +RIDGE & $\Delta$ \\
\midrule
Qwen2.5-VL-3B   & 54.7 & 59.3 & \gain{+4.6} \\
Qwen2.5-VL-7B   & 58.9 & 65.4 & \gain{+6.5} \\
Qwen2.5-VL-32B  & 59.5 & 65.5 & \gain{+6.0} \\
Qwen2.5-VL-72B  & 60.6 & 66.9 & \gain{+6.3} \\
\bottomrule
\end{tabular}
\end{table}

\noindent\textbf{Robustness to Frame Scoring Model.}
RIDGE operates on a precomputed similarity sequence $\{s_t\}$ and is therefore decoupled from the choice of frame scoring model. Table~\ref{tab:scoring_model} compares different scoring backbones. Across LVB, Video-MME, and LVBench, RIDGE consistently improves over uniform sampling under different scoring models. Based on its stable performance, we use BLIP-ITM as the default scoring model in our main experiments.

\noindent\textbf{Robustness to Weight-Generation LLM.}
We test how sensitive RIDGE is to the LLM used for generating the region weights in Table~\ref{tab:llm_weight}. Across Qwen3-3B, Qwen3-8B, and Llama-3.1-8B, the final accuracy changes moderately but remains well above the uniform baseline. This stability is expected: the LLM affects only a compact evidence-preference vector, which modulates the rising/falling extension through coverage intent and then redistributes a fixed frame budget across structural regions.
\begin{table}[t]
\centering
\small
\caption{Sensitivity to the frame--query scoring model across benchmarks. We select BLIP-ITM as the default.}
\label{tab:scoring_model}
\setlength{\tabcolsep}{6pt}
\begin{tabular}{lccc}
\toprule
Scoring Model & LVB & Video-MME & LVBench \\
\midrule
Uniform & 58.9 & 61.2 & 38.5\\
\midrule
CLIP   & 62.0 & 64.3 & 48.9 \\
BLIP-2      & 63.4 & 63.4 & 49.7\\
SigLIP    & 64.7 & 62.5 & 47.8 \\
BLIP-ITM & 65.4 & 63.7 & 50.9 \\
\bottomrule
\end{tabular}
\end{table}

\begin{table}[t]
\centering
\small
\caption{Sensitivity to the weight-generation LLM on LongVideoBench.}
\label{tab:llm_weight}
\setlength{\tabcolsep}{6pt}
\begin{tabular}{lcc}
\toprule
Weight Source & LVB Acc. & $\Delta$ vs Qwen3-8B \\
\midrule
Qwen3-8B (default)       & 65.4 & --- \\
Qwen3-3B                 & 64.2 & -1.2 \\
Llama-3.1-8B             & 64.5 & -0.9 \\
\bottomrule
\end{tabular}
\end{table}

\noindent\textbf{Robustness to Smoothing Scale $\sigma$.}
The Gaussian smoothing scale $\sigma$ controls the granularity of temporal structure recovered from \(\{s_t\}\). A small value leaves noise in the raw similarity curve, while a large value may merge adjacent events. Table~\ref{tab:sigma} sweeps \(\sigma \in \{1.0, 2.0, 3.0, 5.0, 8.0\}\). Accuracy peaks at \(\sigma{=}2.0\) and degrades gracefully when the curve is either under-smoothed or over-smoothed, indicating that RIDGE is not overly sensitive to this hyperparameter.

\begin{table}[t]
\centering
\small
\caption{Sensitivity to the smoothing scale $\sigma$ on LongVideoBench.}
\label{tab:sigma}
\setlength{\tabcolsep}{6pt}
\begin{tabular}{lcc}
\toprule
$\sigma$  & LVB Acc. & $\Delta$ vs Uniform \\
\midrule
1.0       & 63.0 & \gain{+4.1} \\
 2.0   &  65.4 &  \gain{+6.5} \\
3.0       & 63.9 & \gain{+5.0} \\
5.0       & 64.0 & \gain{+5.1} \\
8.0       & 62.8 & \gain{+3.9} \\
\bottomrule
\end{tabular}
\end{table}

Across these axes, RIDGE consistently remains above the corresponding uniform baseline. The gains are therefore not an artifact of a single frame budget, LVLM, scoring model, weight generator, or smoothing setting.

\begin{figure}[t]
    \centering
    \includegraphics[width=1.0\columnwidth]{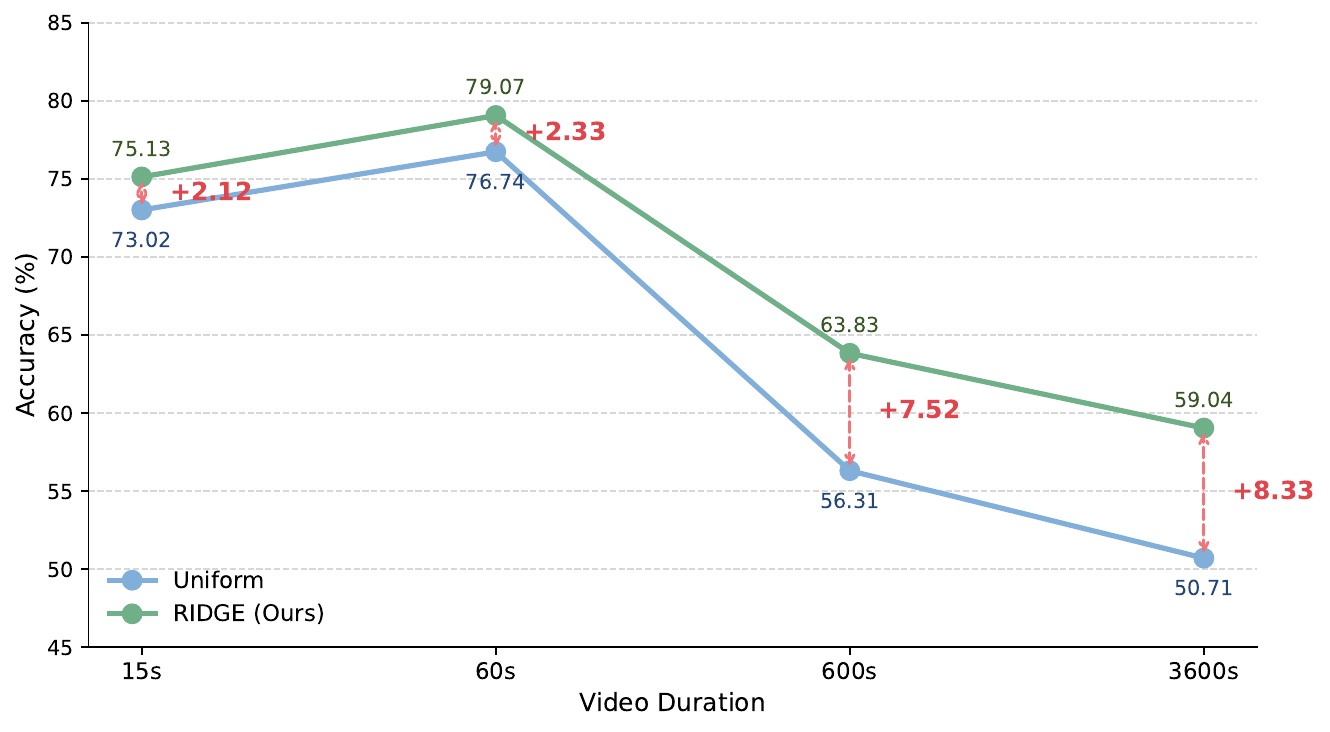}
    \caption{Accuracy on LongVideoBench grouped by video duration.}
    \label{fig:duration_analysis}
\end{figure}

\subsection{Ablation and Analysis}

While Section~4.3 shows that RIDGE is robust to external design choices, this subsection dissects the internal contribution of each component in the final algorithm.

\noindent\textbf{Component Ablation.}
Table~\ref{tab:component} ablates RIDGE at the module level rather than
listing low-level operators. We remove or replace one module from the
full method at a time: temporal phase modeling covers pre-event and
post-event regions; transition modeling covers abrupt-change evidence;
adaptive event-core localization controls event-core spans using local
curve geometry; question-aware evidence weighting controls the
preference vector; and question-adaptive phase extension uses this
preference to modulate how far rising and falling regions extend.

\begin{table}[t]
\centering
\footnotesize
\caption{Component ablation on LongVideoBench. Each row removes or
replaces one component from the full RIDGE pipeline. ``w/o
Question-aware evidence weighting'' uses one fixed weight vector for
all questions, whereas ``w/ Uniform evidence allocation'' assigns equal
weights across evidence types.}
\label{tab:component}
\setlength{\tabcolsep}{5pt}
\resizebox{\columnwidth}{!}{%
\begin{tabular}{lcc}
\toprule
Configuration & Acc. & $\Delta$ \\
\midrule
Uniform sampling & 58.9 & -- \\
RIDGE (full) & \textbf{65.4} & -- \\
\midrule
\multicolumn{3}{l}{\textit{\textbf{Region modeling}}} \\
\quad w/o Temporal phase modeling          & 64.3 & \gain{-1.1} \\
\quad w/o Transition modeling              & 64.2 & \gain{-1.2} \\
\quad w/o Adaptive event-core localization & 64.3 & \gain{-1.1} \\
\midrule
\multicolumn{3}{l}{\textit{\textbf{Budget allocation}}} \\
\quad w/o Question-aware evidence weighting & 63.8 & \gain{-1.6}\\
\quad w/o Question-adaptive phase extension & 64.0 & \gain{-1.4} \\
\quad w/ Uniform evidence allocation        & 64.4 & \gain{-1.0} \\
\midrule
\multicolumn{3}{l}{\textit{\textbf{Selection policy}}} \\
\quad w/o Relevance-based ranking & 64.3 & \gain{-1.1 } \\
\quad w/o Temporal bin selection  & 64.8 & \gain{-0.6} \\

\bottomrule
\end{tabular}%
}
\end{table}

\noindent\textbf{Question-type Analysis.}
A core motivation of RIDGE is to recover evidence distributed across different temporal roles. Figure~\ref{fig:question_type} breaks down LVBench by question type. RIDGE improves substantially on key information retrieval, entity recognition, reasoning, and temporal grounding, all of which often require locating sparse evidence rather than only recognizing the most salient event frame. The gains are smaller on summarization, where broad coverage is helpful but the answer is less tied to a single rising--peak--falling structure.

\noindent\textbf{Video-duration Analysis.}
We further analyze RIDGE on LongVideoBench by grouping videos according
to their duration, as shown in Figure~\ref{fig:duration_analysis}. RIDGE
consistently outperforms uniform sampling across all duration buckets.
The gains are moderate for short videos, with accuracy improving from
73.02 to 75.13 on the 15-second bucket and from 76.74 to 79.07 on the
60-second bucket. As videos become longer, the advantage increases
substantially, reaching +7.52 percentage points on the 600-second bucket
and +8.33 percentage points on the 3600-second bucket. This trend supports the motivation of
RIDGE: uniform sampling can already cover many salient moments in short
clips, while longer videos require preserving more temporally distributed
evidence under the same frame budget.


\section{Conclusion}
\label{sec:conclusion}
We presented RIDGE, a training-free frame selection framework for
long-video understanding. The central idea is to treat the frame-query
similarity sequence as an ordered temporal signal rather than a flat
relevance list. By segmenting this signal into structural regions,
allocating budget according to question-dependent evidence needs, and
selecting frames with region-specific rules, RIDGE preserves not
only salient event cores but also the buildup, aftermath, transitions,
and contextual evidence that are often suppressed by score ranking. Across
four long-video benchmarks and three LVLM backbones, RIDGE delivers
consistent gains over uniform sampling and strong keyframe-selection
baselines, with especially clear benefits in settings that require sparse
or temporally distributed evidence. These results suggest that improving
long-video QA is not only a matter of scoring frames more accurately.

\section*{Limitations}
RIDGE is most useful when the answer depends on evidence distributed
across multiple temporal roles. For short videos, or for questions that
can be answered from a single salient frame, the advantage over uniform
sampling or simple score ranking may be modest because there is less
temporal structure to recover. The method also depends on the quality of
the frame-query scorer and the downstream LVLM. If the scorer does not
produce meaningful variation around the relevant evidence, or if the LVLM
cannot interpret the selected visual evidence, better frame allocation
alone cannot fully solve the question. Finally, RIDGE selects from a
predefined candidate frame sequence. Very brief events, fine-grained OCR,
small objects, or evidence missed by the initial candidate sampling can
remain inaccessible. Future work could combine temporal-shape-aware
selection with denser candidate generation, stronger frame scorers, or
models that jointly refine scoring and reasoning.

\section*{Acknowledgments}
This research was supported by the \emph{National Natural Science
Foundation of China} (No.~62406121) and the \emph{Natural Science
Foundation of Hubei Province, China} (No.~2024AFB189).



\bibliography{custom}

\clearpage

\appendix

\section{Detailed RIDGE Procedure}
\label{sec:appendix_ridge_details}

This section expands the deterministic implementation details omitted
from the main method section. The goal is to make the exact region
definitions and allocation rules reproducible while keeping the main
paper focused on the motivation and high-level design.

\paragraph{Signal preprocessing.}
Given raw similarity scores \(\{s_t\}_{t=1}^{T}\), we first apply
min--max normalization,
\[
    \hat{s}_t = \frac{s_t-s_{\min}}{s_{\max}-s_{\min}},
\]
with the degenerate constant-score case handled by leaving the sequence
unchanged. The normalized signal is smoothed with a one-dimensional
Gaussian kernel,
\[
    \bar{s}_t = (G_{\sigma} * \hat{s})(t),
\]
and the first- and second-order temporal differences are computed as
\(s'_t=\nabla \bar{s}_t\) and \(s''_t=\nabla s'_t\).

\paragraph{Adaptive peak width.}
Peaks are detected on \(\bar{s}_t\) with prominence threshold
\(p_{\min}\) and minimum peak distance of 5. If no peak is detected, the
global maximum is used. For each peak \(p\), we compute its relative
curvature
\[
    r_p = \frac{|s''_p|}{\mathrm{median}(|s''|)+\epsilon}.
\]
The final implementation treats \textsc{peak} as the event core rather
than the whole transition neighborhood. We therefore use an inverse
curvature mapping: sharp peaks receive compact core windows, while
flatter peaks receive wider windows. Specifically, we set
\begingroup
\small
\[
h_p =
\operatorname{clip}\!\left(
  \operatorname{round}\!\left(
    \frac{w_0}{\sqrt{\max(r_p,0.25)}}
  \right),\,2,\,w_{\max}
\right).
\]
\endgroup
This keeps sharp temporal changes around narrow peaks available to the
\textsc{rising}, \textsc{falling}, or \textsc{boundary} labels.

\paragraph{Coverage-modulated region thresholds.}
From the question-preference vector, we compute
\[
    c =
    \frac{w_{\mathrm{rise}} + w_{\mathrm{fall}} + w_{\mathrm{context}}}
    {w_{\mathrm{peak}} + w_{\mathrm{rise}} +
     w_{\mathrm{fall}} + w_{\mathrm{context}} + \epsilon}.
\]
The rising/falling slope threshold is then
\[
    \tau_s =
    \left(k_s(1-c) + 0.1c\right)\cdot \mathrm{std}(s'),
\]
and the maximum one-sided extension is
\[
    L_{\max} =
    \min\!\left(
      \mathrm{round}\!\left(k_e\max(\sigma,1)(1+c)\right),\,
      \left\lfloor T/4 \right\rfloor
    \right).
\]
Starting from each peak boundary, we extend left while \(s'_t>\tau_s\)
and right while \(s'_t<-\tau_s\), up to \(L_{\max}\). For boundary
detection, we only consider frames that are still labeled as background
after peak and slope-region assignment. Let this residual set be
\(\mathcal{B}\). A frame \(t\in\mathcal{B}\) is labeled as boundary if
\[
    |s'_t| >
    \mathrm{median}(|s'|_{\mathcal{B}})
    + k_b \cdot \mathrm{std}(|s'|_{\mathcal{B}}).
\]

\paragraph{Question-preference mapping.}
The six question weights are
\[
\begin{aligned}
\mathbf{w}=(&w_{\mathrm{peak}}, w_{\mathrm{slope}},
w_{\mathrm{rise}},\\
&w_{\mathrm{fall}}, w_{\mathrm{boundary}},
w_{\mathrm{context}}).
\end{aligned}
\]
They are mapped to five effective region weights. The generic slope
weight represents a preference for frames where query relevance changes
quickly; in the five-region partition, this evidence is represented by
\textsc{boundary}, so the slope weight is added to the boundary weight:
\[
\begin{aligned}
W_{\mathrm{peak}} &= w_{\mathrm{peak}}, \\
W_{\mathrm{rising}} &= w_{\mathrm{rise}},\\
W_{\mathrm{falling}} &= w_{\mathrm{fall}},\\
W_{\mathrm{boundary}} &= w_{\mathrm{boundary}} + w_{\mathrm{slope}},\\
W_{\mathrm{background}} &= w_{\mathrm{context}}.
\end{aligned}
\]
Each region receives
\[
    n_r =
    \mathrm{round}\!\left(
      N \frac{W_r}{\sum_{r'}W_{r'}}
    \right),
\]
capped by its available number of frames. Any remaining budget is
redistributed to available regions in descending effective-weight order.

\paragraph{Budget-aware sub-bin selection.}
For \textsc{rising} and \textsc{falling}, sub-bin selection is used only
when both the allocated budget and the region length are at least 3. The
region indices are sorted by time and split into \(n_r\) approximately
equal bins, where \(n_r\) is the region budget. From each bin, RIDGE
selects the frame with the largest \(\bar{s}_t\). If \(n_r<3\) or the
region contains fewer than 3 frames, the method falls back to ranked
selection by \(\bar{s}_t\).

\paragraph{Default hyperparameters.}
Unless otherwise stated, we use \(\sigma=2.0\),
\(p_{\min}=0.15\), \(w_0=3\), \(w_{\max}=10\), \(k_s=0.5\),
\(k_e=4.0\), \(k_b=2.0\), and a minimum temporal gap of 2 for ranked
selection regions.

\section{Weight-generation Prompt}
\label{sec:appendix_weight_prompt}
When RIDGE uses question-specific region weights, the auxiliary LLM
receives only the video question and the instruction below. It returns
a six-dimensional preference vector in JSON, which the allocation step
parses deterministically.

\begin{lstlisting}[style=ridgeprompt]
You are an expert in video analysis. Given a video question,
analyze what visual evidence is needed to answer it, and then
output a weight vector that controls how frames are selected
from the video's query-relevance curve.

Video question:
{question}

Weight dimensions:
- peak_similarity: Are frames highly similar to the question
  needed? (High for descriptive questions, e.g.,
  "What is shown in the scene?")
- slope_abs: Are frames with rapid changes needed?
  (High for dynamic or procedural questions.)
- rising_slope: Are frames from an increasing similarity trend
  needed? (High for causal or setup questions, e.g., "Why...?")
- falling_slope: Are frames from a decreasing similarity trend
  needed? (High for follow-up or consequence questions, e.g.,
  "What happens afterward?")
- boundary_change: Are event-boundary or abrupt-change frames
  needed? (High for questions about turning points.)
- context_density: Are frames from information-dense regions
  needed? (High for questions requiring context.)

Return strictly in the following JSON format. Each weight
must be an integer from 0 to 10:
{
  "peak_similarity": <int>,
  "slope_abs": <int>,
  "rising_slope": <int>,
  "falling_slope": <int>,
  "boundary_change": <int>,
  "context_density": <int>,
  "reasoning": "<brief rationale>"
}

Output JSON only. Do not output any other content.
\end{lstlisting}

\section{Compared Frame-selection Methods}
\label{sec:appendix_compared_methods}

\paragraph{Uniform sampling.}
Uniform sampling is the non-query-aware reference in the main
comparison. It places a fixed frame budget at evenly spaced temporal
locations and therefore preserves coarse coverage without requiring a
scoring model or an additional selection algorithm. Its limitation is
that the same temporal allocation is used for every question: sparse
query-relevant evidence may fall between samples, while visually
uninformative spans still consume frames.

\paragraph{AKS.}
Adaptive Keyframe Sampling (AKS) is a plug-in keyframe selector designed
to trade off prompt relevance against temporal coverage
\citep{tang2025adaptive}. AKS obtains prompt--frame matching scores from
a lightweight vision--language model and approximates coverage with a
recursive temporal binning objective. Its adaptive procedure contrasts
two extreme policies---top-score sampling, which can collapse into a
narrow relevant interval, and binned sampling, which enforces coverage
more strongly---and recursively decides whether a temporal bin should be
split or should directly return its top-scoring frames. The selected set
therefore remains question-aware while avoiding a purely peak-centered
allocation when coverage is beneficial.

\paragraph{BOLT.}
BOLT studies training-free frame selection for off-the-shelf long-video
VLMs and adopts query-guided inverse transform sampling as its strongest
selection strategy \citep{liu2025bolt}. It first builds a frame
probability distribution from query--frame similarity scores, with a
sharpness parameter controlling how strongly the distribution favors high
scores. Sampling from the inverse cumulative distribution gives highly
relevant frames a larger selection probability while retaining nonzero
probability for lower-score frames. This design is intended to reduce the
redundancy of deterministic top-\(k\) selection and retain more temporal
context and diversity at inference time without training the downstream
VLM.

\paragraph{MDP$^3$.}
MDP$^3$ formulates frame selection as a list-wise problem that should jointly
respect query relevance, diversity among selected frames, and temporal
sequentiality \citep{sun2025mdp3}. It constructs a query-conditioned
similarity matrix through a conditional Gaussian kernel in an RKHS and
uses a determinantal point process to favor relevant yet non-redundant
frame subsets. To preserve sequential structure, the video is segmented
and the DPP selection for each segment is conditioned on the previous
segment. The resulting selection-size allocation is modeled as a Markov
decision process and solved with dynamic programming, yielding a
training-free selector that explicitly discourages list-wise redundancy.

\paragraph{FOCUS.}
FOCUS is a training-free, model-agnostic keyframe selector that treats
short temporal clips as arms in a budgeted combinatorial
pure-exploration problem \citep{zhu2025focus}. Each clip maintains an
empirical relevance estimate and a Bernstein-style confidence radius, so
the method can spend scoring budget on clips that are either promising or
uncertain. Its practical two-stage schedule first performs coarse
parallel exploration over clips and then applies batched optimistic
exploration to the strongest candidates before choosing clips by their
empirical means. Within selected clips, FOCUS constructs frame-level
selection distributions from observed and interpolated relevance rewards
to produce the final keyframes under the token budget.

\section{Evaluation Benchmarks}
\label{sec:appendix_benchmarks}

\paragraph{Video-MME.}
Video-MME is a broad benchmark for evaluating multimodal LLMs on video
analysis \citep{fu2025video}. It contains 900 manually curated videos
and 2{,}700 multiple-choice question--answer pairs across six visual
domains and 30 fine-grained categories. The videos span short, medium,
and long regimes from 11 seconds to one hour, and the benchmark also
provides subtitles and audio to probe multi-modal evidence use. This
coverage makes Video-MME useful for testing frame selection across both
compact clips and longer temporal contexts; in our experiments, we use
the video-only setting without subtitles.

\paragraph{LongVideoBench.}
LongVideoBench evaluates long-context interleaved video--language
understanding on 3{,}763 web videos with subtitles and 6{,}678
human-annotated multiple-choice questions \citep{wu2024longvideobench}.
Its central task is referring reasoning: a referring query identifies one
or more relevant video contexts, and the model must retrieve and reason
over details in those contexts to answer the question. The benchmark
separates perception questions from relation questions and further
organizes them into 17 categories over videos ranging from seconds to
hour-long inputs. It is therefore sensitive to whether a selector keeps
the specific frames needed for detailed long-context reasoning.

\paragraph{MLVU.}
MLVU is a multi-task benchmark built to diagnose long-video
understanding across video genres, durations, and task formats
\citep{zhou2025mlvu}. Its videos span roughly three minutes to more than
two hours with an average duration of about 12 minutes, covering
real-world sources such as movies, documentaries, TV series, egocentric
videos, and surveillance footage as well as animated and game videos.
MLVU defines nine task categories, including topic reasoning, anomaly
recognition, summarization, needle QA, ego reasoning, plot QA,
sub-scene captioning, action counting, and action ordering. These tasks
mix holistic, single-detail, and multi-detail demands, making the
benchmark informative for selectors that must balance global coverage
with sparse local evidence.

\paragraph{LVBench.}
LVBench targets extreme long-form video understanding
\citep{wang2025lvbench}. Its 103 publicly sourced videos are selected
from long videos with coherent structure, multiple chronological events,
and content that can be understood visually without relying on audio;
the benchmark contains 1{,}549 manually annotated QA pairs with an
average video duration exceeding one hour. Its questions probe six core
capabilities: key information retrieval, event understanding, entity
recognition, reasoning, temporal grounding, and summarization. Because
the evaluated videos are long and multi-event, LVBench exposes whether a
frame selector preserves evidence needed for long-horizon retrieval and
temporal reasoning rather than only salient isolated moments.

\section{Complexity and Runtime Analysis}

RIDGE introduces only post-processing on a precomputed similarity
sequence and requires no iterative LVLM calls. Stage~1 (smoothing, finite
differences, peak detection, and region labeling) runs in
\(\mathcal{O}(T)\). Stage~2 operates over a fixed set of structural
regions. Stage~3 is dominated by ranked selection and backfilling, giving
a worst-case complexity of \(\mathcal{O}(T \log T)\). This is
asymptotically comparable to a plain top-\(k\) selector and cheaper than
LVLM-iterative selectors such as A.I.R., which require additional model
calls.

Table~\ref{tab:time} reports wall-clock selection time per video on LVB,
excluding the shared cost of computing \(s_t\). RIDGE adds only
8.1\,ms of post-processing, remains close to lightweight non-iterative
selectors, and is far faster than LVLM-iterative methods. The optional Qwen3-8B weight call is performed once per input question at inference time, adding a small question-side preprocessing cost; however,
it does not introduce additional downstream LVLM calls.

\begin{table}[t]
\centering
\small
\caption{Per-video selection latency on LongVideoBench at $N{=}32$.
Time excludes the shared $\{s_t\}$ computation.
$^{\S}$ denotes LVLM-iterative methods.}
\label{tab:time}

\setlength{\tabcolsep}{2pt}
\begin{tabular}{@{}lcc@{}}
\toprule
Method
& \makecell{Selection Time}
& \makecell{LVLM Calls} \\
\midrule
Uniform
& $<$ 1\,ms
& 0 \\

AKS
& 5.2\,ms
& 0 \\

BOLT
& 12.4\,ms
& 0 \\
A.I.R.$^{\S}$
& 3.4\,s
& $\geq 1$ \\
\makecell[l]{RIDGE (w/o LLM)}
& 8.1\,ms
& 0 \\
\makecell[l]{RIDGE (w/ LLM)}
& 8.1\,ms + 0.6\,s$^{*}$
& 0 \\
\bottomrule
\end{tabular}

\par\smallskip
\raggedright\footnotesize
$^{*}$ Auxiliary LLM call is performed once per input question at
inference time.
\end{table}

\section{Hyperparameter Sensitivity}
\label{sec:appendix_hyperparam}

Beyond the smoothing scale $\sigma$ analyzed in
Section~\ref{sec:robustness}, the region segmentation stage of RIDGE
contains four core hyperparameters: the peak prominence threshold
$p_{\min}$, which determines how many local maxima are retained as event
cores; the base peak half-width $w_0$, which specifies the anchor size
of the curvature-adaptive peak window; the slope threshold coefficient
$k_s$, which controls the temporal extent of the rising and
falling regions around each event core; and the boundary threshold
coefficient $k_b$, which affects the detection of abrupt changes
among residual background frames.

Table~\ref{tab:hyperparam_sensitivity} reports the sensitivity of
RIDGE to these hyperparameters on LVB and MLVU. For each sweep, one
hyperparameter is varied while the remaining three are fixed to their
default values. Across all tested configurations, RIDGE remains
substantially above the uniform sampling baseline, which achieves
58.9\% on LVB and 57.3\% on MLVU. This indicates that the performance
gain of RIDGE is not tied to a narrow hyperparameter setting.

Among the four hyperparameters, $k_s$ exhibits the largest
variation, with ranges of 1.4 percentage points on LVB and 1.1 percentage points on MLVU. This is
expected, as $k_s$ directly determines how much temporal context
is included around each detected event core. In contrast, $k_b$
varies by 0.7 percentage points on LVB and 1.1 percentage points on MLVU. This
suggests that boundary frames play a useful but secondary role, since
they usually occupy only a small fraction of the sampling budget.
The peak prominence threshold $p_{\min}$ and the base half-width $w_0$
show moderate sensitivity. Very small $p_{\min}$ values may introduce
spurious event cores, whereas overly large values may suppress
weaker but informative events. Similarly, too small or too large
values of $w_0$ can under-cover or over-cover local event regions.
Overall, the default setting provides the best average performance
while maintaining stable behavior under reasonable perturbations.

We additionally evaluate two safety-clamp parameters: the maximum
peak half-width $w_{\max}{=}10$ and the maximum extension factor
$k_e{=}4.0$. These parameters are rarely activated in
practice. Varying $w_{\max}$ within $[6,15]$ and $k_e$ within
$[2.0,6.0]$ changes accuracy by less than 0.5 percentage points on
both benchmarks; therefore, we omit the detailed sweep.

\begin{table}[t]
\centering
\small
\caption{Hyperparameter sensitivity of RIDGE on LVB and MLVU with
Qwen2.5-VL-7B ($N{=}32$). Accuracy (\%) is reported. The default
configuration is highlighted. Bold entries in the first four columns
indicate the hyperparameter being varied.}
\label{tab:hyperparam_sensitivity}
\setlength{\tabcolsep}{6pt}
\renewcommand{\arraystretch}{1.08}
\begin{tabular*}{0.95\linewidth}{@{\extracolsep{\fill}}cccccc@{}}
\toprule
$p_{\min}$ & $w_0$ & $k_s$ & $k_b$ & LVB & MLVU \\
\midrule
0.15 & 3 & 0.5 & 2.0 & 65.4 & 69.6 \\
\midrule
\textbf{0.05} & 3 & 0.5 & 2.0 & 65.1 & 69.2 \\
\textbf{0.10} & 3 & 0.5 & 2.0 & 64.8 & 69.3 \\
\textbf{0.30} & 3 & 0.5 & 2.0 & 64.4 & 69.7 \\
\midrule
0.15 & \textbf{1} & 0.5 & 2.0 & 64.6 & 69.0 \\
0.15 & \textbf{5} & 0.5 & 2.0 & 65.1 & 69.8 \\
0.15 & \textbf{8} & 0.5 & 2.0 & 64.2 & 68.6 \\
\midrule
0.15 & 3 & \textbf{0.3} & 2.0 & 64.5 & 69.8 \\
0.15 & 3 & \textbf{0.8} & 2.0 & 64.9 & 69.2 \\
0.15 & 3 & \textbf{1.0} & 2.0 & 64.0 & 68.7 \\
\midrule
0.15 & 3 & 0.5 & \textbf{1.5} & 64.7 & 70.1 \\
0.15 & 3 & 0.5 & \textbf{2.5} & 65.2 & 69.4 \\
0.15 & 3 & 0.5 & \textbf{3.0} & 64.8 & 69.0 \\
\bottomrule
\end{tabular*}
\end{table}

\section{Multi-Peak Analysis}
\label{sec:appendix_multi_peak}

RIDGE keeps multiple candidate peaks rather than restricting selection
to the global maximum. We analyze this behavior on LongVideoBench with
Qwen2.5-VL-7B and $N{=}32$, comparing with uniform sampling and AKS,
the strongest competing selector on this benchmark in
Table~\ref{tab:main_results}. We partition instances according to whether
RIDGE's default peak detector returns one or multiple peaks.

\begin{table*}[t]
\centering
\small
\caption{Performance on single-peak and multi-peak LongVideoBench
subsets with Qwen2.5-VL-7B and $N{=}32$. Accuracy (\%) is reported.}
\label{tab:multi_peak}
\setlength{\tabcolsep}{10pt}
\begin{tabular}{lccccc}
\toprule
Subset & Uniform & AKS & RIDGE & $\Delta$ vs. Uniform & $\Delta$ vs. AKS \\
\midrule
All         & 58.94 & 63.20 & 65.37 & +6.43 & +2.17 \\
Single-peak & 71.51 & 75.11 & 75.50 & +3.99 & +0.39 \\
Multi-peak  & 54.46 & 58.96 & 61.76 & +7.30 & +2.80 \\
\bottomrule
\end{tabular}
\end{table*}

RIDGE improves over AKS by 2.80 percentage points on the multi-peak
subset, compared with 0.39 percentage points on the single-peak subset.
This larger gain is consistent with RIDGE retaining evidence beyond the
dominant peak. On the non-short multi-peak subset, an average of 7.15\%
of RIDGE-selected frames fall inside the strongest-peak window. Thus,
the selections are not concentrated in that window, although RIDGE does
not guarantee that every detected peak receives a selected frame.

\section{Held-out Hyperparameter Control}
\label{sec:appendix_heldout}

To assess whether RIDGE depends on evaluation-set tuning, we create a
fixed, video-disjoint development/evaluation split of LongVideoBench.
Using only the 20\% development subset, we select among 16 pre-declared
configurations, freeze the selected configuration, and evaluate it on the
disjoint 80\% LongVideoBench subset and the other three benchmarks without
further adaptation. The paper-default configuration was fixed before all
sensitivity sweeps and is evaluated under the same protocol.

\begin{table}[t!]
\centering
\small
\caption{Held-out hyperparameter control with Qwen2.5-VL-7B and
$N{=}32$. Accuracy (\%) is reported. Avg. excludes LVB-dev and averages
the four evaluation sets.}
\label{tab:heldout_control}
\setlength{\tabcolsep}{4pt}
\renewcommand{\arraystretch}{1.12}
\begin{tabular*}{\columnwidth}{@{\extracolsep{\fill}}lccc@{}}
\toprule
Benchmark & Uniform & \makecell{Default} & \makecell{Held-out} \\
\midrule
LVB-dev      & 57.7 & 68.2 & 69.4 \\
LVB-held-out & 59.3 & 64.7 & 64.2 \\
Video-MME    & 61.2 & 63.7 & 64.3 \\
MLVU         & 59.7 & 69.6 & 69.4 \\
LVBench      & 38.5 & 50.9 & 51.4 \\
Avg.         & 54.7 & 62.2 & 62.3 \\
\bottomrule
\end{tabular*}
\end{table}

The configuration selected on LVB-dev uses $\sigma{=}2.0$,
$p_{\min}{=}0.15$, $w_0{=}3$, $k_s{=}0.5$, $k_e{=}4.0$, and
$k_b{=}2.5$; the safety clamps $w_{\max}{=}10$ and the minimum temporal
gap of 2 remain fixed. It achieves essentially the same four-benchmark
average as the paper default (62.3 vs. 62.2) and consistently outperforms
uniform sampling across all four evaluation sets. These results show that
RIDGE's gains are preserved when its hyperparameters are selected without
access to the reported evaluation labels.

\section{More Visualization Results}

\label{sec:more_visualization_results}

\begin{figure*}[p!]
    \centering

    \includegraphics[
        height=0.42\textheight,
        width=\textwidth,
        keepaspectratio
    ]{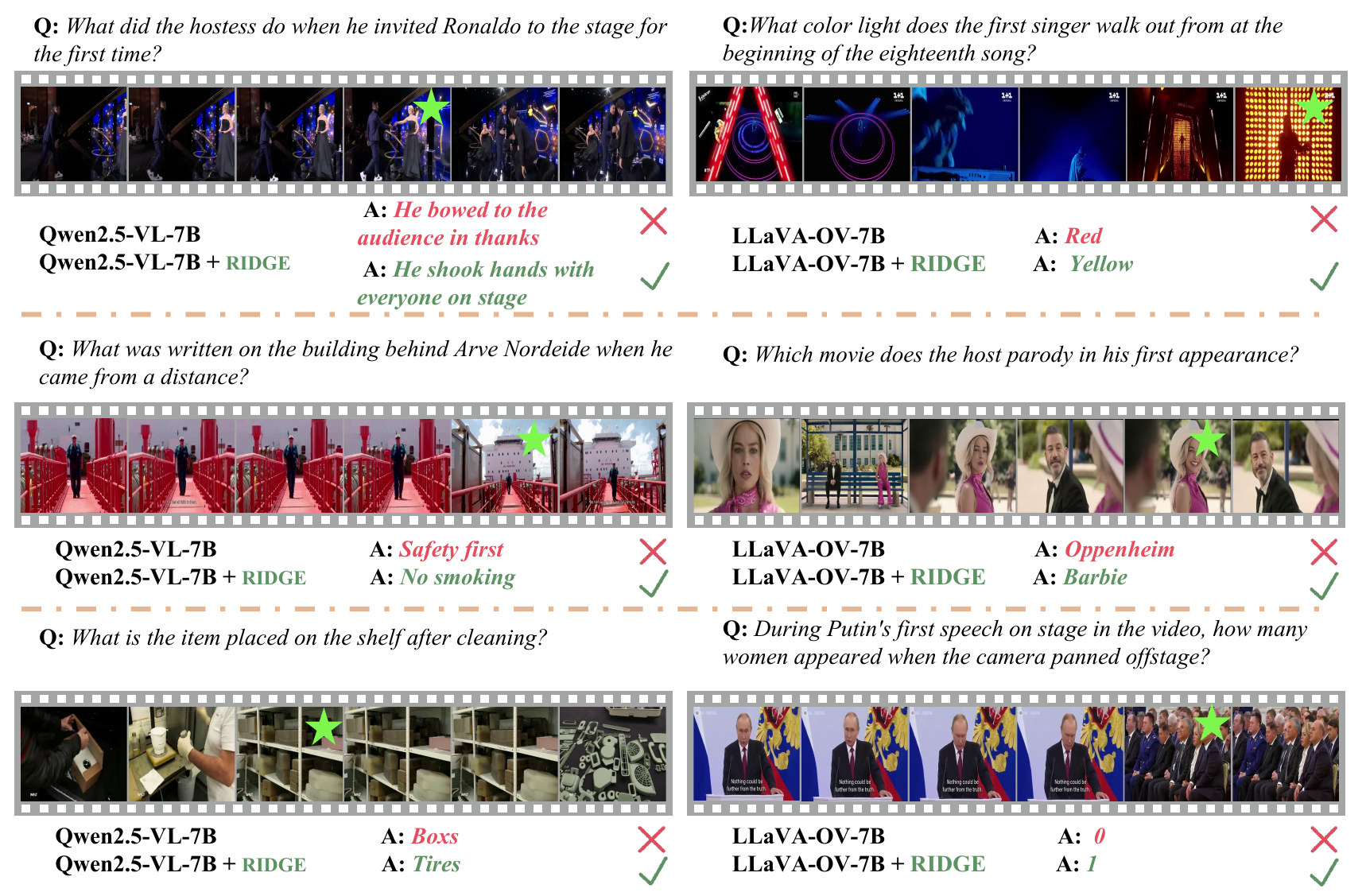}

    \caption{
Qualitative examples showing how RIDGE improves long-video question
answering with Qwen2.5-VL-7B and LLaVA-OV-7B. Each panel presents a
question, sampled video frames, and the answers produced with and
without RIDGE. Green stars mark key frames selected by RIDGE, while
red crosses and green check marks denote incorrect and correct answers,
respectively. Without RIDGE, the models may miss fine-grained visual
evidence and produce plausible but incorrect answers. RIDGE shifts
frame selection toward query-relevant moments, helping the models
recover the evidence needed to answer the questions correctly.
    }
    \label{fig:case_1}

    \vspace{2mm}

    \includegraphics[
        height=0.30\textheight,
        width=\textwidth,
        keepaspectratio
    ]{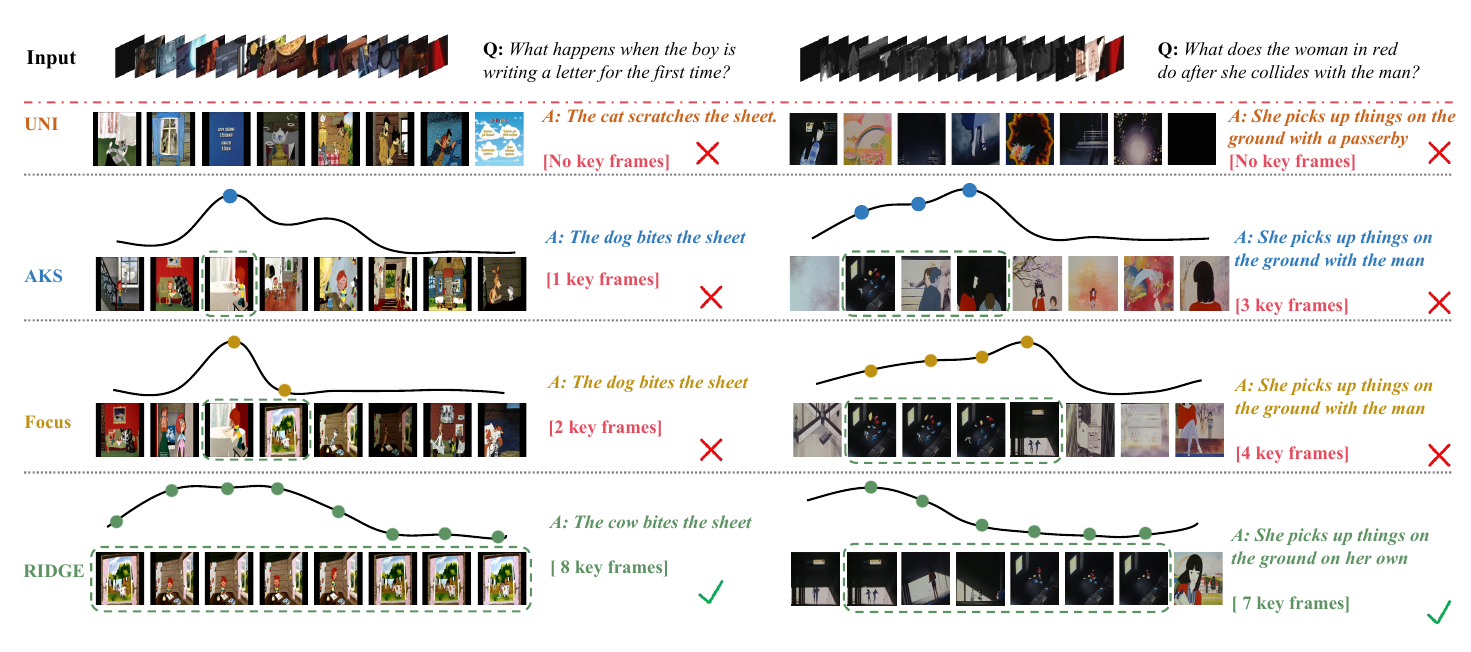}

    \caption{
Qualitative comparison between uniform sampling, AKS, FOCUS, and RIDGE
on long-video question answering. The top row shows the input video frames
and the query. For each method, the dashed green boxes indicate the selected
key frames or selected temporal spans, and the curve shows the query-frame
relevance trajectory. Red crosses denote incorrect answers and green check
marks denote correct answers. Compared with prior keyframe selection methods,
RIDGE explicitly uses the shape of the relevance curve to preserve event cores,
pre-event buildup, aftermath, and transition evidence, which leads to more
faithful visual grounding.
    }
    \label{fig:case_2}
\end{figure*}

Figure~\ref{fig:case_1} presents six representative examples spanning
diverse question types—action recognition, text reading, object
identification, color discrimination, movie reference, and
counting—to illustrate the broad applicability of RIDGE. In each case
the base model selects frames that miss the decisive visual evidence,
leading to plausible but incorrect answers. 

For instance, Qwen2.5-VL-7B
misreads ``No smoking'' as ``Safety first'' and mistakes tires for boxes,
while LLaVA-OV-7B confuses the parodied movie and miscounts the number
of women offstage. After integrating RIDGE, the key-frame selection
shifts to the query-relevant moments (marked by green stars), and all
six answers are corrected. These results suggest that RIDGE generalizes across the evaluated backbones and improves performance across diverse visual reasoning skills.

Figure~\ref{fig:case_2} compares RIDGE with uniform sampling,
AKS~\cite{tang2025adaptive}, and FOCUS~\cite{zhu2025focus}, providing
additional qualitative evidence for the effectiveness of RIDGE. The two
examples illustrate a common failure mode in long-video question answering:
the visually decisive evidence is not always contained in a single
high-relevance frame. Instead, answering the question often requires a short
temporal process, including what happens before the event, the event core
itself, and what happens afterward.

\end{document}